# A Systematic Study and Analysis of Bengali Folklore with Natural Language Processing Systems

**Mustain Billah[1*], Mostafijur Rahman Akhond[2], Md. Mynoddin[3], Md. Nasim Adnan[4], Syed Md. Galib[5], Rizwanur Rahad[6], M Nurujjaman Khan[7]**

[1,3,4,5,6,7] Department of Computer Science and Engineering,
Jashore University of Science and Technology, Jashore, Bangladesh
[2]Department of Computer Science and Engineering,
Rangamati Science & Technology University, Rangamati, Bangladesh
[*]Corr. Email: mu.billah@just.edu.bd

## Abstract

Folklore, a solid branch of folk literature, is the hallmark of any nation or any society. Such as oral tradition; as proverbs or jokes, it also includes material culture as well as traditional folk beliefs, and various customs. Bengali folklore is as rich in-depth as it is amazing. Nevertheless, in the womb of time, it is determined to sustain its existence. Therefore, our aim in this study is to make our rich folklore more comprehensible to everyone in a more sophisticated computational way. Some studies concluded various aspects of the Bengali language with NLP. Our proposed model is to be specific for Bengali folklore. Technically, it will be the first step towards Bengali natural language processing for studying and analyzing the folklore of Bengal.

## I. Introduction

Bengali folklore is a strong and prosperous branch of Bengali literature. Folklore is a collection of legends, proverbs, myths, riddles, superstitions, and rituals that have been handed down from generation to generation. The folklore of Bangladesh can disclose a lot about the social and ethnic history of the country and attitudes of the people or the values, customs, and logic of the previous inhabitants. There is a considerable amount of Bengali folklore in the collection of Bengali Folk literature. Bengali folk literature survives despite being developed by illiterate communities and passed down orally from generation to generation. Individual folklore has evolved into a communal representation of customs, emotions, opinions, and values. The Mouryas, Guptas, Palas, Senas, and Muslims fought for dominance of the region in the early 3rd century. As a result, the inhabitants were influenced by their way of lifestyle and cultural features. Then ships from Portugal, France, and England docked at Bengal harbor. They have left behind not only their belongings but also their heritage. Each nationality has left not only a physical but also a cultural imprint, which was then combined to form the basis of contemporary culture. Since folklore is primarily oral, it often involves memory skills with language and stylistic patterns. Bengali folklore includes legends, proverbs, myths, riddles, superstitions, and rituals, all of which exist in different forms throughout the Bengali community, whether educated or not. Several nations have made a significant impact on Bangladeshi folklore. As a result, Bangladeshi folklore has several features that can be partially explained by historical conditions. Puthis are antique books manufactured in rural Bangladesh that contain folk tales and religious legends. These writings were read aloud in public by educated individuals for both entertainment and education. The Puthis were written in Bangla and Songskrito by Munshis. The word English Folk Tales refers to almost all kinds of folk tales. In Bengali it is called "রুপকথা বা পরীকাহিনি" but in English, it is known as fairy tales. Fairy tales are based on the story of fairy's attributes like beauty, magic, highness, and so on. But fairy may not remain in all fairy tales. Demonstration of heroism by the prince or emperor by going to the unfamiliar city of "রাক্ষস-খোক্কস', impossible accomplishments, rescue and marriage of princesses or princesses of different countries, magical powers, divine help, etc. are the features of the 'রূপকথা'. In German, such a story is called Marchen. In Bengal, DakshinaranjanMitraMajumdar edited 'ঠাকুরমার ঝুলি', 'ঠাকুরদাদার ঝুলি', 'থানদিদির থলে' etc., a collection of such stories.

There is another kind of long folktale, the events of which are as incredible and thrilling as fairy tales but often find geographical and historical similarities with the events and the names of the characters. For example, Arabic novels, Persian novels, stories of বিক্রমাদিত্য or তালবেতাল, etc. King Solomon or Caliph Harun-al-Rashid of Arabic fiction and বিক্রমাদিত্য of পঞ্চ-বেতাল-বিংশতি are historical figures. These long stories are called Novella in English and "রোমাঞ্চকথা' in Bengali. The mixture of fairy tales and thrillers is noticeable in the stories like 'সয়ফুলমুলুক-বদিউজ্জামাল' narrated in Bengali books. Stories based on animals are also called 'উপকথা'. And all the animal stories that have been told advice or proverbs are called "fables" in English. And in Bengali, it is said "নীতিকথা'. 'সংস্কৃত পঞ্চতন্ত্র ও হিতোপদেশ' are excellent examples of such fables. In English which may say ' Anecdote ". The instructive story of people like preceptors, gods, prophets, monks, saints, saints, etc. which is said 'উপদেশকথা'. In English is said "রাঙ্গাকথা' or 'হাসিরগল্প' is also included in Folk Tales. Jasimuddin's 'বাঙ্গালির হাসির গল্প', Upendrakishore Roy Chowdhury's 'টুনটুনির বই', Ashraf Siddiqui's 'Toontoony and Other tales' are excellent examples.

Some stories happened once and gained popularity through word of mouth. These are called local stories (স্থানীয়কাহিনি). For example, in Subodh Ghosh's ' কিংবদন্তির দেশে', Ashraf Siddiqui's 'কিংবদন্তির বাংলা', East Bengal-lyricist 'কমলাসুন্দরী', 'চৌধুরীরলড়াই' etc., There were such stories but they were not collected much.

Other types of folk tales such as 'বাঘের গায়ে ডোরাকাটা কেন', 'কুটুম পাখির গায়ের রঙ হলুদ কেন' or a particular tree, river, why and how the constellation was created, etc., are told locally as 'সৃষ্টিকথা'; In English, it is called Etiological tale or Nature sage.

There is a kind of story in Bengali which is called 'গাঁজাখুরি গল্প' (Tales of lying in English). Such as: 'পি-পুফি-শু' (পিঠ পোড়ে, ফিরে শোও), 'দুইজোয়ানের গল্প' (those who bite with palm trees and banyan trees, dry the water of the pond with one kiss)(যারা তালগাছ ও বটগাছ দিয়ে দাঁতন করে, এক চুমুকে পুকুরের পানি শুকিয়ে ফেলে) etc. There are also puzzle stories 'ধাঁধামূলক গল্প'; This type of story was written to test the intelligence of the new son-in-law, the new bride, or the job seekers of the royal court. Patterns of these kinds of stories are found in 'গোপাল ভাঁড়'.

There is a kind of story that revolves around one event after another in the form of rhyme, which is known as 'শিকলি গল্প' (Chain Tale). Tuntuni's book - 'টুনটুনি ও নাপিত', 'নাপিত ও শৃগাল (নাক কেটে নরুন পেলাম)etc. are such kinds of stories.

In Bengali folktales, characters such as unhappy ducks, poor crows, chatty Tuntuni birds, foolish tigers, and animals and birds that express themselves like humans are all common. It exemplifies the hamlet residents' ingenuity and innovation. The poor crow who jumped into the fire to make a sparrow for his companion is a parable that laments societal class disparities. The story's moral is to be happy with what you have. These tales are 'true' not because they took place, but because they frequently contain 'truth' or 'wisdom'.

Nonetheless, it is determined to continue to exist in the womb of time. There has been a lot of research into folk language and folklore in other languages, but it is unlikely that it will be found in Bengal. For this, we want to use advanced counting procedures from Bengali folklore to give this oral culture more stability and richness. Many researchers have offered various natural language processing-based solutions to various challenges. NLP has been used in a few studies to highlight different elements of the Bengali language. Our proposed model should be unique to Bengali folklore; it should be a part of it. We'll create a morphological representation of folklore here. Here we are going to build a morphological representation of folklore. It will be the first step toward Bengali natural language processing for the study and analysis of Bengali folk literature. Although, it is nothing but a sequential model with some hybrid dense layer empowered by some core techniques of natural language processing which will be covered in detail in sections III and IV.

## II. Related Works

Bengali folklore is a well-established and thriving area of Bengali literature. Even though the fact that a study on several branches of the Bengali language has been done, it is believed that little work has been done on Bengali folklore. Although work on folk literature and folk stories has been done in a variety of languages, it is unlikely to be found in Bengali. As a result, we want to use Bengali folklore computational approaches to make this oral folklore more stable and rich. Many researchers have proposed many Natural Language Processing-based solutions to solve some challenges. In article [4] developed a technique for extracting information about literary characters from unstructured texts using natural language processing and domain ontology reasoning. There are three primary phases in the suggested method. (I) Identify the key characters and the portions of the story where these characters are described or act, (II) Demonstrate the system in a folktale domain scenario, and (III) The system depends on a folktale ontology that they developed based on Propp's model for folktale morphology.

Daniel Suciu and Adrian Groza[5] presented an approach for detecting literary characters from folktales, in which the extraction of characters from free texts is driven by an ontology that stores folktale domain knowledge. In the paper, they showed the way for ontological interactions with GATE's Natural Language Processing to improve character recognition accuracy, and they tested the approach against a variety of folktales.In article [6] developed a process for generating a summary for storytelling. As it's difficult to tell a long story to children. If the story gets summarized only then that's easy for them to understand. For that, they proposed a summarization process using the word frequency concept. Their process is based on five major steps: (I) Eliminate special characters (II) Tokenization to get a set of words (III) Remove stop word components and count word frequency (IV) Find the word frequency that the sentences are scored (V) Combine them into a single paragraph.In article [9],the thesis will analyze different approaches in ontology deployment from a basic lexical ontology based on WordNet to an extended model for ontology development involving different domains of knowledge in the field of Folklore.In article [10], the authors suggested a knowledge independent system is trained using supervised learning methods and applied to the rest of the corpus using classifiers such as the Naïve Bayes, K-nearest Neighbor, Support Vector Machine (SVM) and others.In article [11], large volumes of ICH data were gathered for this investigation, and domain knowledge was retrieved from the text data using Natural Language Processing (NLP) technology. The knowledge graph was constructed after building a knowledge base based on domain ontology and instances for Chinese intangible cultural assets. The ICH knowledge graph was used to depict the pattern and characteristics of intangible cultural heritage. They proposed the intangible cultural heritage knowledge graph to assist knowledge management and provide a service to the public.In article [12], the authors developed the DID system that when applied to children stories starts by classifying the utterances that belong to the narrator (indirect discourse) and those belonging to the characters taking part in the story (direct discourse). In article [13], present a technique for assigning particular narrative roles to characters in stories in this work. To do so, they propose combining natural language processing techniques with domain knowledge gleaned from Propp's folktale morphology. A matrix, which encapsulates the narrative domain knowledge, is used to express the linkages between character roles.In article [14], they propose a narratological grounded definition of character for discussion at the workshop, and demonstrate a preliminary yet straightforward supervised machine-learning model with a small set of features that performs well on two corpora.

In article [15], propose a method while most natural language understanding systems rely on a pipeline-based architecture, certain human text interpretation methods are based on a cyclic process between the whole text and its parts: the hermeneutic circle.

## II. Proposed Methodology

Methodology applied in this research work is illustrated in Figure 1. Raw string data is fed into the system as input. After following different steps, folk theme based morphological meaning is generated as the output.

### *A. Problem Statement*

We deal with the problem by identifying all nominal characters representing folk characters in the input text and classifying them according to the folktale they belong to.

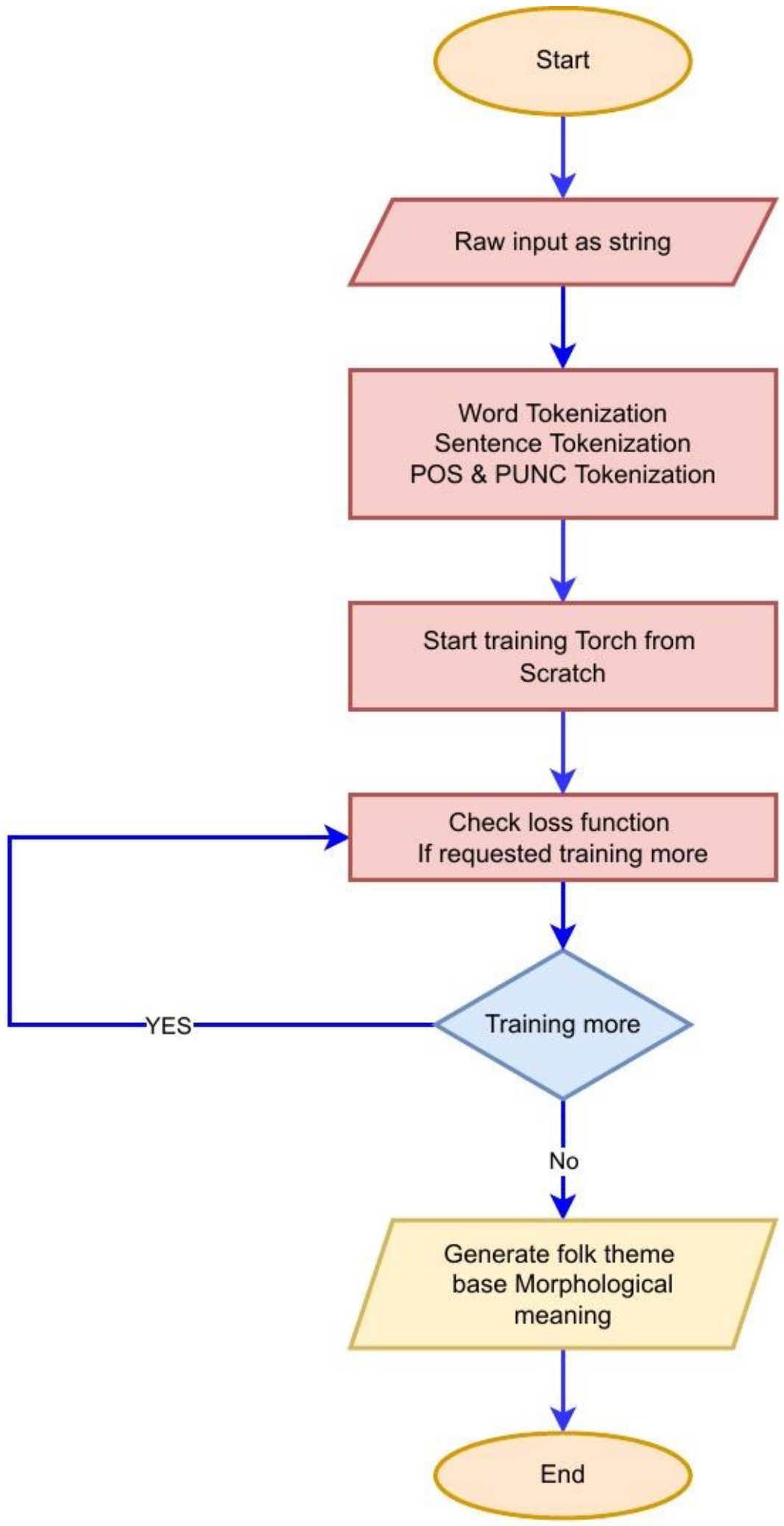

Figure 1: Flow chart of the proposed methodology

Characters in folktales are often defined by their roles or tasks rather than their names, which makes this conundrum even more intriguing. In addition, it has taken on the role of a different character in a different section of the folktale, yet the system must recognize all of these instances as representing the same figure. In other words, we require a component that lets us encode the logic that governs the characters' actuality. Describing our solution's logic is its cornerstone.

### B. Problem Analysis

***Tokenizers:*** FolkBangla NLP provides three different tokenization algorithms for Bengali text. As rule-based tokenizers, FOLKBANGLA NLP provides Basic Tokenizer, a punctuation splitting tokenizer, and NLTK tokenizer. We modified the nltk tokenize output to use the NLTK tokenizer for Bengali, taking into account the changes in punctuation between English and Bengali. Under model-based tokenization, FOLKBANGLA NLP provides a sentencepiece tokenizer for Bengali text called Bengali Sentence piece. The Bengali sentence piece API includes both a pre-trained sentencepiece model and a training sentencepiece model.

***Embeddings:*** FOLKBANGLA NLP has two embedding alternatives for Bengali words: word2vec (Mikolov et al., 2013) and fasttext (Bojanowski et al., 2016). Bengali word2vec and fasttext have two options: one is to use a pre-trained model to incorporate Bengali words, and the other is to train Bengali word2vec or fasttext from scratch. For the bot embedding model, we have used the gensim5 embedding API and trained on Bengali corpus.

For fulfill our challenges, firstly, we want to make an input of raw text of any folktales of Bangladesh. There are some models that exist like BERT [1]; we tokenized the word and sentence of Bengali folktales input. When the tokenization was completed, we just clean up the text from punctuation and took the parts of speech [2] tagged, then make an LSTM - Deep learning model with a learning rate of 0.1. This produced a summary of the folktales and kept the morphological meaning. For our working purpose we have used the tensorflow to build the final folklore model.

Though it is working well, it was not exactly what we wanted to build. So we decided to create a new model for specifically folk literature. We have named it the FolkBangla model.

### C. The FolkBangla Model

Before diving into technical details we’ve to analyze some key factors of folk tales. There is some proven method for studying Folklore of any language. One is the Propp's model, originally proposed by Vladimir Proppand it was for studying vast soviet folklore but later the thirty-one rules described any language or origin folktales should be applied to the rules. While studying Bengali folklore for natural language processing we interestingly found the Propp's model. The 31 rules [3] of the method of identifying character and narrative finely fit with Bangla folklore too. The morphological meaning of every folklore is not an outsider of Propp's model. By the theorem, Vladimir Propp proposed a common model for all folklore of the world. His approach was applied to all genres of narratives, including folklore, literature, film, television series, theater, games, mimes, cartoon strips, ads, dance forms, sports comments, film theory, news reporting, story generation, and interactive drama systems, among others. His approach was applied to all genres of narratives, including folklore, literature, film, television series, theater, games, mimes, cartoon strips, ads, dance forms, sports comments, film theory, news reporting, story generation, and interactive drama systems, among others. Let’s consider a linear equation,

$$a = bx + cy \ldots\ldots\ldots\ldots\ldots\ldots. (1)$$

Here, a is a set of bengali words, working on two dimensional vector space to obtain loss function (gradient) to check performance. Here x,y are stochastic, while b , c are non-stochastic , pre-defined weight variables.

We built a model according to the Propp’smodel to identify folk characters. This is the first part of studying Bengali folklore. We have gone through some other work. We have found [7] the paper says Improvements in statistical computation frequently lead to improvements in statistical modeling. It is uncommon, for example, for an old model to be re-parameterized solely for computational reasons, but this new configuration then inspires a new family of models useful in applied statistics. One reason this phenomenon may have gone overlooked in statistics is that re-parameterizations do not change the likelihood. In a Bayesian framework, a rearrangement of parameters, on the other hand, frequently implies a new family of prior distributions. For statistical computation of folklore they suggested a statistical model. However we're using Propp's model to determine character.

Now, '□□□□', '□□□□', '□□□□□□□□' with these words we will test while model training to see how the model is learning. Dimensions in embedding vectors. 200 scalar values will be stored against one word. Embedding Size is set to 200. If there are more words than this in our corpus, we will exclude them. Here, Maximum Vocabulary Size is set to 50000. In this process, we will omit words that are too few times in our data.Minimum Occurrence is set to 10.

When we compile the data, we will also use some words from our left and right sites and the number of data we want to make by going to the same wordHere, Skip Window is set to 3 and Number Skips is set to 2. After that, we like to use desired negative data when calculating loss. First, loss was 23% and finally, after training the loss is 5%.

We have used sampled numbers about 64. After running the codes, we have found Unique Characters. In the input, document Kiran Mala, Total Words: 2726.

[' ', '!', '"', '""', '*', ',', '-', '.', ':', ';', '?', '।', 'ঁ', 'ং', 'ঃ', 'অ', 'আ', 'ই', 'ঈ', 'উ', 'এ', 'ঐ', 'ও', 'ক', 'খ', 'গ', 'ঘ', 'ঙ', 'চ', 'ছ', 'জ', 'ঝ', 'ঞ', 'ট', 'ঠ', 'ড', 'ঢ', 'ণ', 'ত', 'থ', 'দ', 'ধ', 'ন', 'প', 'ফ', 'ব', 'ভ', 'ম', 'য', 'র', 'ল', 'শ', 'ষ', 'স', 'হ', 'া', 'ি', 'ী', 'ু', 'ূ', 'ৃ', 'ে', 'ৈ', 'ো', 'ৌ', '্', 'ৎ', 'ড়', 'য়', '০', '১', '২', '৩', '৪', '৫', '৬', '৭', '৮', '৯']

Figure 2: Found Unique Characters

Finally, we have the most common folk words and the number of their occurrences.

```
print("Words count:", len(text_words))
print("Unique words:", len(set(text_words)))
print("Vocabulary size:", VOCABULARY_SIZE)
print("Most common words:", dictionary[:20])

data        : [6, 0, 11, 1, 9, 11, 0, 0, 1, 0, 0, 0, 0, 0, 0, 2, 0, 0, 0, 0, 0, 0, 0, 0, 2, 0, 0, 1, 0, 0, 0, 0, 0, 0, 0, 
Totla data  : 2726
--------------------------------------------------
data in text: ['কিরণমালা', 'অজ্ঞাত লোককাহিনি আশ্রিত শব্দ', 'এক', 'রাজা', 'আর', 'এক', 'অজ্ঞাত লোককাহিনি আশ্রিত শব্দ', 'অজ্ঞা
--------------------------------------------------
Words count: 2726
Unique words: 1251
Vocabulary size: 41
Most common words: [('অজ্ঞাত লোককাহিনি আশ্রিত শব্দ', 2050), ('রাজা', 56), ('কি', 33), ('না', 33), ('বরুণ', 30), ('করিয়া', 28)
```

Figure 3: Most common folk words and the number of their occurrences

## IV. Models and Data set

### *A. Pre-trained Model*

BanglaFolk NLP provides several pre-trained models such as: i)Sentencepieceii) Word2vec iii)Fasttext

**Sentencepiece:** To train multiple language models, we need a better word embedding level vocabulary. To build subword-based vocabulary, we train a sentencepice model on Bengali Kiran Mala dataset.

**Word2vec:** We trained a Bengali word2vec model on Bengali Kiran Mala dataset using the gensim word2vec pipeline. The embedding dimension of 200, the window size of 3, the minimum number of word occurrences of 10 has been used to train our word2vec model. We train it for a total of 25000 iterations.

**Fasttext:** To train the BengaliFasttext model, we used the Kiran Mala dataset in Bengali. To train fasttext, we have utilized the following parameters: embedding dimension 200, windows size 3, number of minimum word occurrences 10, model type skip-gram. After 100 epochs of training, our loss is 0.2718668.

**SpaCy:** spaCy is an open-source software library designed in the Python and Cython programming languages for advanced natural language processing. Unlike the widely used NLTK, spaCy is geared on developing production-ready software. SpaCy also supports deep learning, allowing users to combine statistical models generated with popular machine learning libraries such as TensorFlow, PyTorch, and MXNet with spaCy's native machine learning library Thinc.

### *B. Data Set*

For training sentencepiece, word2vec, fasttext we used Bengali raw folklore text data from Dakshinaranjan Mitra Majumdar edited 'ঠাকুরমারঝুলি'; specially 'Kiran Mala' folk story. In the data set, total words are 2726.

## V. Result Analysis

We have tested our dataset and model at various parameters. By the comparison, We have found that our model is working sharp with better precision, recall and F1 scores. We have plotted the precision scores and the plot is showing a clear advantage of the FolkBangla model for studying specifically folk studies.

Table 1: Scoring Comparison

| Model Name | Precision | F1 | Recall |
|---|---|---|---|
| SpaCy | 62.67 | 62.117 | 59.78 |
| Fasttext | 71.45 | 69.76 | 75.88 |
| FolkBangla | 88.56 | 85.38 | .85.8 |

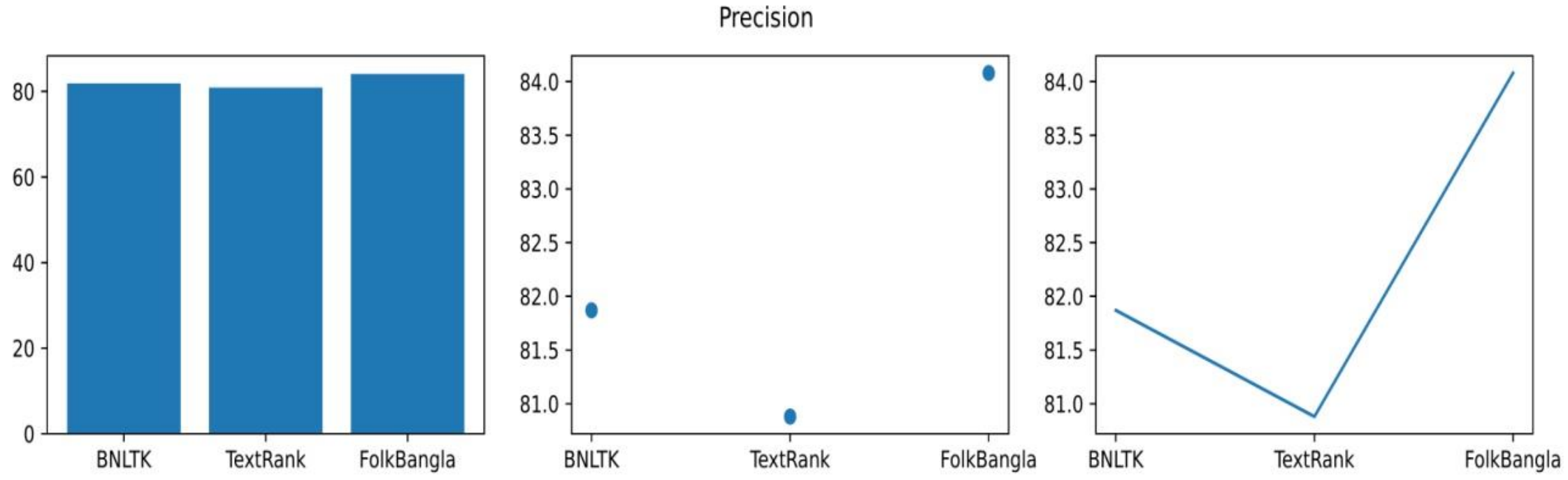

Figure 4: Comparison of Precision for different models for Folklore

This is the initial model and currently doing character identification according to Proop's model and giving summary of specific folk stories. But it also does fine in other literature. We are considering BNLTK (Bangla Natural Language Processing Toolkit),TextRank and FolkBangla on Bengali news summarization.

When we plot the precision of other literature, we have seen our model is working better on other Bengali texts too. And this happened because we trained our model using the Tensorflow API from scratch.

Table 2: Comparison on other Bengali literature

| Model Name | Precision | F1 | Recall |
|---|---|---|---|
| BNLTK | 81.87 | 79.91 | 80.54 |
| TextRank | 80.88 | 77.65 | 79.38 |
| Folk Bangla | 84.08 | 81.38 | 82.12 |

Figure 5 is indicating performance on other type of bengali written text. It is very likely to see FolkBangla performing well. A sample output of Proposed Method is Shown in Figure 6.

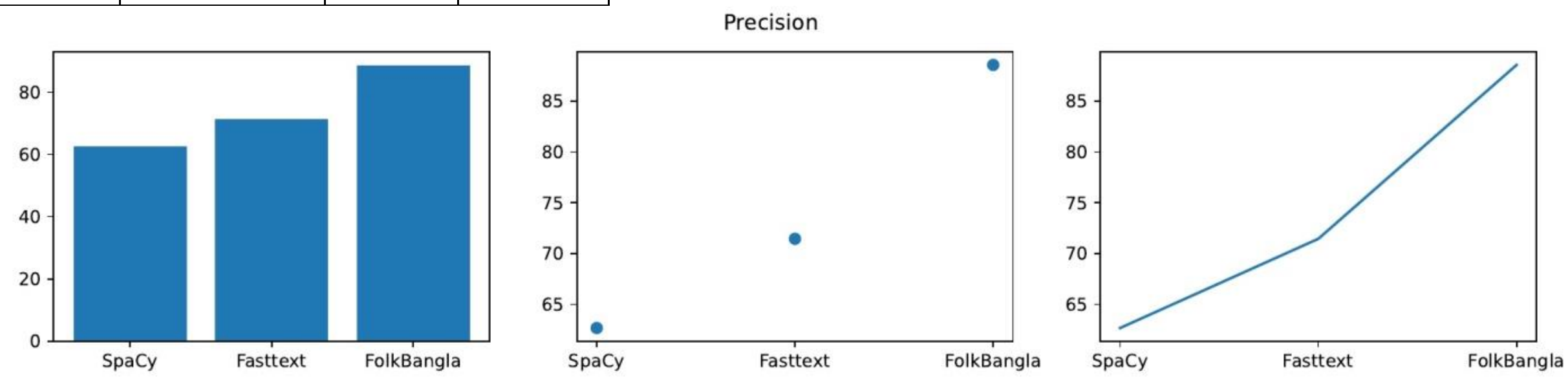

Figure 5: Precision scores of different models for news summarization

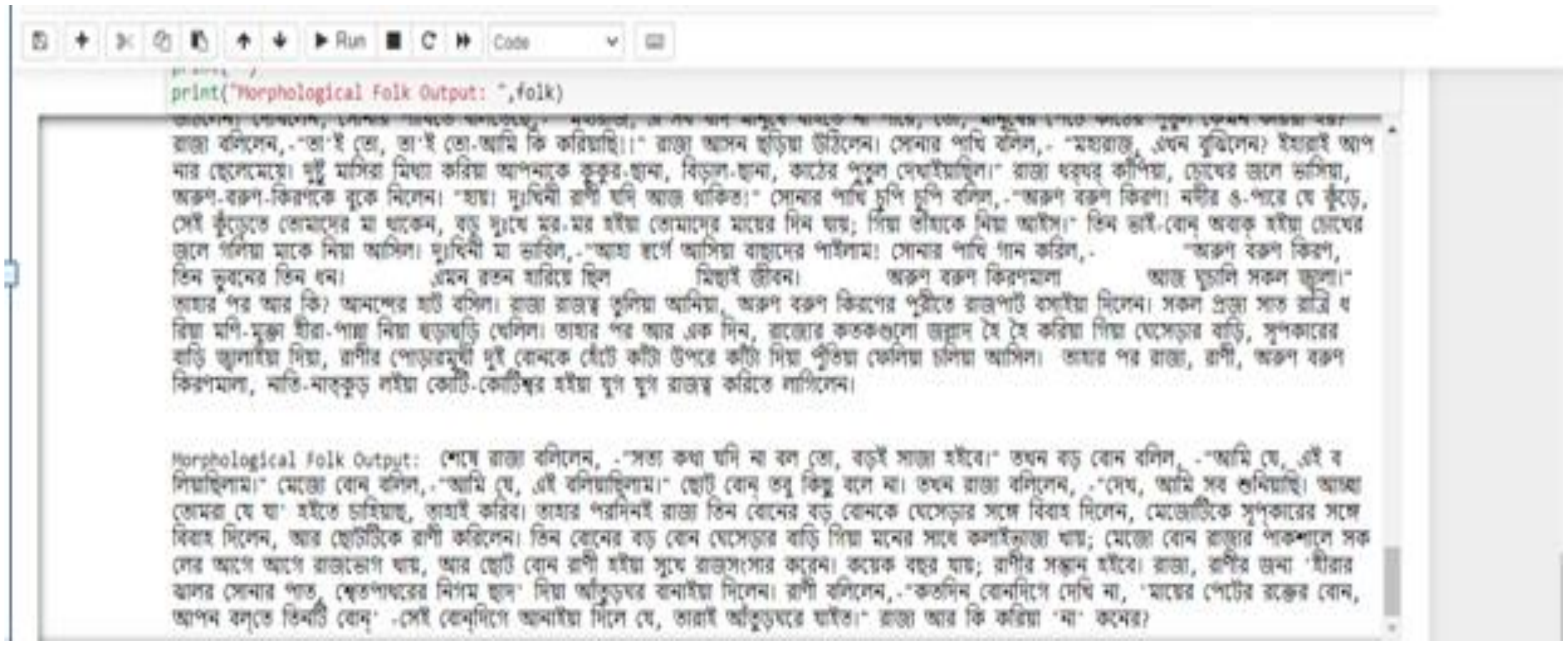

Figure 6: A Sample Output

## VI. Conclusion and Future work

Folklore in Bengali literature is a well-established and vibrant field. Even though the fact that numerous branches of the Bengali language have been studied, little work on Bengali folklore is considered to have been done. Folk literature and folk storytelling have been studied in a variety of languages, but Bengali is unlikely to be among them. As a result, we have applied computational Bengali folklore methodologies to make this traditional folklore more sustainable. This is the initial step toward researching Bengali folklore using natural language processing and computer science. Next we will move on a regional domicile based approach. It is somewhat area based folk studies with follicle tung based stories or literature.  This discussion will continue by digging deeper into Bengali folk literature. Our Motto is 'Folklore is our flowing heritage.